\documentclass[preprint,12pt]{elsarticle}

\usepackage{amssymb}
\usepackage{amsmath}
\usepackage{url}
\usepackage{makecell}
\usepackage{multirow}
\usepackage{tabularx} % for 'tabularx' env. and 'X' col. type
\usepackage{ragged2e} % for \RaggedRight macro
\usepackage{booktabs} 
\usepackage{cleveref}
\usepackage{stfloats}
\usepackage{subfig}

\def\figurecrefname{Fig.}

\crefformat{figure}{#2\figurecrefname~#1#3}

\def\tablecrefname{Tab.}

\crefformat{table}{#2\tablecrefname~#1#3}

\journal{Image and Vision Computing}

\usepackage{caption}
\begin{document}

\begin{frontmatter}



\title{PSTF-AttControl: Per-Subject-Tuning-Free Personalized Image Generation with Controllable Face Attributes}

\cortext[cor1]{Corresponding author}

\author[oo1,oo2]{Xiang liu*} %% Author name
\ead{liux750@chinaunicom.cn}
\author[oo1,oo2]{Zhaoxiang Liu*}
\ead{liuzx178@chinaunicom.cn}
\author[oo1,oo2]{Huan Hu}
\author[oo1,oo2]{Zipeng Wang}
\author[oo1,oo2]{Ping Chen}
\author[oo1,oo2]{Zezhou Chen}
\author[oo1,oo2]{Kai Wang}
\author[oo1,oo2]{Shiguo Lian*}
\ead{liansg@chinaunicom.cn}

\affiliation[oo1]{organization={Unicom Data Intelligence},%Department and Organization
            addressline={China Unicom}, 
            city={Beijing},
            postcode={100013}, 
            country={P.R.China}}

\affiliation[oo2]{organization={Data Science \& Artificial Intelligence Research Institute},%Department and Organization
            addressline={China Unicom}, 
            city={Beijing},
            postcode={100013}, 
            country={P.R.China}}

%% Abstract

\begin{abstract}
Recent advancements in personalized image generation have significantly improved facial identity preservation, particularly in fields such as entertainment and social media. However, existing methods still struggle to achieve precise control over facial attributes in a per-subject-tuning-free (PSTF) way. Tuning-based techniques like PreciseControl have shown promise by providing fine-grained control over facial features, but they often require extensive technical expertise and additional training data, limiting their accessibility. In contrast, PSTF approaches simplify the process by enabling image generation from a single facial input, but they lack precise control over facial attributes. In this paper, we introduce a novel, PSTF method that enables both precise control over facial attributes and high-fidelity preservation of facial identity. Our approach utilizes a face recognition model to extract facial identity features, which are then mapped into the \(W^+\) latent space of StyleGAN2 using the e4e encoder. We further enhance the model with a Triplet-Decoupled Cross-Attention module, which integrates facial identity, attribute features, and text embeddings into the UNet architecture, ensuring clean separation of identity and attribute information. Trained on the FFHQ dataset, our method allows for the generation of personalized images with fine-grained control over facial attributes, while without requiring additional fine-tuning or training data for individual identities. We demonstrate that our approach successfully balances personalization with precise facial attribute control, offering a more efficient and user-friendly solution for high-quality, adaptable facial image synthesis. The code is publicly available at https://github.com/UnicomAI/PSTF-AttControl.

\end{abstract}

% \centering
% \includegraphics[width=1.0\textwidth]{images/graph.png}
% \vspace{-0.2cm}

% \begin{highlights}
% \item A per-subject-tuning-free method is proposed for personalized image generation with precise control over facial attributes.
% \item An attribute-controlled synthesis strategy is used for data augmentation on the FFHQ dataset, enabling controllable facial attribute learning.
% \item A Triplet-Decoupled Cross-Attention module integrates identity, attribute, and text features into UNet while preserving identity consistency.
% \end{highlights}

%% Keywords
\begin{keyword}


Personalized generation, Controllable face attributes
\end{keyword}

\end{frontmatter}



\section{Introduction}
Personalized image generation with high-fidelity facial identity preservation has developed rapidly in recent years, driven by applications in fields like entertainment and social media. However, existing methods still fall short of achieving two critical goals simultaneously: precise control over facial attribute generation and a per-subject-tuning-free (PSTF) approach. Here, ‘per-subject-tuning-free’ refers to methods that do not require fine-tuning for each new identity, although they may involve a one-time, global training phase for their components. Achieving these objectives is essential for creating realistic, adaptable, and accessible facial image generation models.

Tuning-based methods \cite{gal2022image,ruiz2023dreambooth,hu2021lora,kumari2023multi,wei2023elite}, such as PreciseControl \cite{parihar2025precisecontrol}, have shown promise by utilizing the \(W^+\) latent space in StyleGAN2. This space enables fine-grained control over facial attributes, allowing users to perform detailed edits such as subtle adjustments to attributes and expressions. Despite its strengths, the tuning-based approach has notable drawbacks: it often requires technical expertise to fine-tune the model parameters, demands a set of training images for each identity, and involves time-consuming processes. These factors make tuning-based methods less practical for broader, user-friendly applications.

On the other hand, PSTF methods \cite{valevski2023face0,peng2024portraitbooth,li2024photomaker,chen2023dreamidentity,ye2023ip,xiao2024fastcomposer,wang2024instantid,guo2024pulid} offer significant advantages by allowing users to generate personalized images with only a single facial input image, removing the need for parameter adjustments. These methods typically leverage large datasets of face images and identity adapters to embed the facial identity, making them accessible and efficient. However, they often lack the ability to control facial attributes precisely, which limits their versatility in generating nuanced, highly customized images.

In this work, we introduce a novel \textbf{p}er-\textbf{s}ubject-\textbf{t}uning-\textbf{f}ree approach, PSTF-AttControl, that enables precise \textbf{control} over facial \textbf{att}ribute generation while maintaining high-fidelity facial identity. Our method extracts facial identity information using a face recognition model, capturing the unique features of the input face. We then use the e4e encoder for StyleGan2 \cite{tov2021designing} to map the facial input image to the \(W^+\) latent space of StyleGAN2 \cite{viazovetskyi2020stylegan2}. Next, we integrate facial identity features, facial attribute features, and text embeddings into the UNet architecture through a Triplet-Decoupled Cross-Attention module. After training on the FFHQ dataset \cite{karras2019style}, our model is able to generate personalized images that preserve facial identity with just a single input image. Additionally, by modifying the facial attribute components in the \(W^+\) space, we enable personalized generation with fine-grained control over facial attributes. 

Our contributions can be summarized as follows:
\begin{itemize}
    \item Precise Control and PSTF Generation: We propose an approach that achieves both precise control over facial attribute generation and a PSTF process.
    \item Data Augmentation with Attribute-controlled Synthesis: Using the FFHQ dataset, we employ an attribute-controlled synthesis approach for data augmentation, enabling the model to learn controllable facial attribute.
    \item Triplet-Decoupled Cross-Attention: This module effectively integrates identity features, attribute features, and text embeddings into the UNet architecture, ensuring that the attribute features do not interfere with the identity features.
\end{itemize}

\section{Related Work}

\subsection{Personalized image generation with facial Identity}
\textbf{Text-to-Image generation with Diffusion Models.} 
Text-to-image diffusion models \cite{nichol2021glide, ramesh2022hierarchical, saharia2022photorealistic,rombach2022high, balaji2022ediff, huang2023composer, wang2025landmarks, zhou2024enhancing, verma2025novel}, trained on vast datasets of internet-scale image-text pairs, achieve high-quality image generation with impressive generalization capabilities. Models such as Stable Diffusion are built upon the latent diffusion model \cite{rombach2022high}, which processes images within a latent representation space rather than directly in pixel space. This approach allows high-resolution image generation with improved efficiency by reducing computational demands typically associated with diffusion models. With pretrained language models such as CLIP \cite{radford2021learning} and T5 \cite{raffel2020exploring} transforming text into embeddings that are integrated seamlessly into the diffusion model, these models use text conditions to control image content generation. This embedding-based conditioning improves the coherence and precision of generated images in alignment with the text prompts. 

\textbf{Tuning-based Personalized image generation with facial Identity.} Personalized image generation in text-to-image models aims to enable pretrained models to produce images that align with descriptive prompts while maintaining consistency with the facial identity in reference images. Textual Inversion \cite{gal2022image} presents an innovative approach to this challenge by leveraging the embedding space of a frozen text-to-image model to introduce new pseudo-words that represent specific concepts or objects. This method allows for the creation of personalized images through natural language instructions, offering a high degree of flexibility and control over the generation process. DreamBooth \cite{ruiz2023dreambooth} builds upon these concepts by introducing a fine-tuning process that enables the model to associate a unique identifier with a specific subject, allowing for the generation of novel and diverse images of that subject across various contexts while preserving key visual features. This approach is particularly powerful as it requires only a few images of the subject to achieve this personalization, making it highly accessible for users with limited data. Celeb-basis \cite{yuan2023inserting} constructs a compact basis from celebrity embeddings, enabling the integration of new identities into diffusion models with just a single photo and minimal parameters. This approach offers efficient personalization, maintaining the model's original capabilities while allowing new identities to interact with existing concepts.

Tuning-based methods enable personalized image generation through fine-tuning, allowing models to adapt to new identities with a handful of data. However, these methods require users to manage complex training processes and are less efficient for inference, as fine-tuning is necessary for each new identity, making them less convenient for rapid, on-demand image generation.

\textbf{PSTF Personalized image generation with facial Identity.} PSTF approaches \cite{valevski2023face0,peng2024portraitbooth,li2024photomaker,chen2023dreamidentity,ye2023ip,xiao2024fastcomposer,li2024stylegan,wang2024instantid,guo2024pulid} enable users to generate personalized images with only a single facial input image, removing the need for parameter adjustments. IP-Adapter \cite{ye2023ip} offers a PSTF solution that is compatible with pre-trained text-to-image diffusion models. It achieves this by introducing a lightweight adapter with a decoupled cross-attention mechanism, which allows the model to leverage image prompts without the need for extensive fine-tuning. This approach is significant because it maintains the original capabilities of the base model while enhancing its flexibility to incorporate facial identity from a single image prompt. W-Plus-Adapter \cite{li2024stylegan} incorporates StyleGAN’s editable \(W^+\) space into the SD model, enabling identity customization. However, integrating \(W^+\) into SD presents a key challenge: the conversion of real images into \(W^+\) vectors in StyleGAN often results in a loss of detail, compromising identity preservation. Despite leveraging a substantial number of training pairs \(\{I_f, W^+\}\) to establish a robust mapping, limitations remain in maintaining fine-grained identity features. InstantID \cite{wang2024instantid} leverages a decoupled cross-attention mechanism and integrates ControlNet \cite{controlnet} to enhance zero-shot identity-preserving generation. The decoupled cross-attention refines facial identity retention by selectively attending to identity-related features, while ControlNet enables better control over image features to balance facial resemblance and image quality. This combination advances previous PSTF methods by achieving detailed, identity-consistent results without fine-tuning, using only a single image reference. PuLID \cite{guo2024pulid} introduces a PSTF approach for identity customization in generative models, focusing on speed and fidelity. By employing contrastive alignment, it preserves identity features across images without fine-tuning. The method achieves high-quality personalization with minimal latency, leveraging a robust contrastive framework to enhance similarity to the reference image while maintaining generation efficiency. PuLID’s design makes it particularly useful for applications requiring rapid, high-fidelity facial generation based on a single input.

These PSTF methods enable personalized image generation without fine-tuning, using a single input image. However, they primarily rely on text prompts to control facial attributes, limiting fine-grained adjustments. For example, they cannot smoothly modify specific features like the degree of a smile, making them less flexible for nuanced customization compared to tuning-based methods.

\subsection{Personalized image generation with Controllable Face Attribute}

PreciseControl \cite{parihar2025precisecontrol} introduces a novel approach that integrates two categories of models by conditioning a text-to-image model with the \(W^+\) space from StyleGAN2, using pre-trained StyleGAN2 encoders. By manipulating the latent vectors in \(W^+\), PreciseControl enables more precise control over facial attributes, offering a finer level of customization.

However, when modifying the latent vectors in \(W^+\), the images generated by StyleGAN2 may not maintain consistent facial identity with the original reference image. This means that PreciseControl may not guarantee high identity consistency in generated images. Additionally, since PreciseControl is a tuning-based method, it requires fine-tuning during inference, making it less convenient for on-demand image generation.

\section{Proposed Method}

\label{method}

\subsection{Preliminaries}

\textbf{Text-to-Image Diffusion Models.} In this work, we leverage Stable Diffusion XL (SDXL) as our foundational text-to-image model, a state-of-the-art variant in latent diffusion models. SDXL operates within a compressed latent space using a pre-trained variational autoencoder (VAE), enabling a more computationally efficient and scalable generation process by reducing the dimensional complexity of the data.  

The training of SDXL consists of two stages. First, a VAE encodes high-dimensional image data into a low-dimensional latent space while preserving both global structure and fine-grained details. Then, a diffusion model is trained within this latent space, conditioned on text embeddings produced by a frozen pre-trained language model, typically CLIP, which effectively captures the semantic content of text prompts.  

SDXL’s architecture is centered on a modified U-Net with attention layers that enhance contextual awareness during denoising. By integrating cross-attention mechanisms, the model accurately aligns text and image features, ensuring semantic coherence throughout the generation process. Additionally, we employ classifier-free guidance, a technique that balances the influence of conditional and unconditional models during sampling, allowing for fine-tuned control over image quality and adherence to prompts.

\textbf{Encoder for StyleGAN}: Style-based GANs have been widely a popular choice for generating realistic, object-specific images, particularly faces, due to their disentangled latent space, which supports versatile image editing. To edit face images effectively, these models require an accurate inversion of the input image into the \(W^+\) latent space. High-quality inversion is crucial for enabling precise, fine-grained control over facial attributes. For this purpose, we utilize the e4e encoder, pretrained on the FFHQ dataset, as our face attribute encoder to map facial images into \(W^+\) space. In this space, we apply attribute adjustments through directional changes, denoted as $\Delta W$, to modify specific facial attributes.

\textbf{Image Prompt Adapter}: IP-Adapter enables a pretrained text-to-image diffusion model to integrate both text and image prompts seamlessly. Unlike existing adapters that often fail to fully capture image details within pretrained architectures, the IP-Adapter employs a decoupled cross-attention mechanism. This design adds dedicated cross-attention layers for image features without modifying the existing text-focused layers, preserving the original model structure while enhancing image feature embedding. In particular, a frozen CLIP image encoder is used to extract global image embeddings, which are then projected into a sequence of feature vectors that match the text feature dimensions. The decoupled cross-attention mechanism then embeds these image features by combining separate cross-attention outputs for text and image. This approach retains the original cross-attention layers for text while adding new cross-attention layers exclusively for image prompts, ensuring that image and text information are both effectively leveraged in generating high-quality, prompt-based images.

Building on the IP-Adapter approach, InstantID enhances it by using a face recognition model to extract identity features, which are then integrated into the UNet using a decoupled cross-attention mechanism. This ensures better preservation of identity during image generation. Additionally, InstantID introduces a ControlNet module that incorporates spatial information, enabling fine-grained control over both identity and text features, while maintaining consistency with the original UNet settings.

\subsection{Methodology}

\begin{figure*}[t]
  \centering
   \includegraphics[width=1\linewidth]{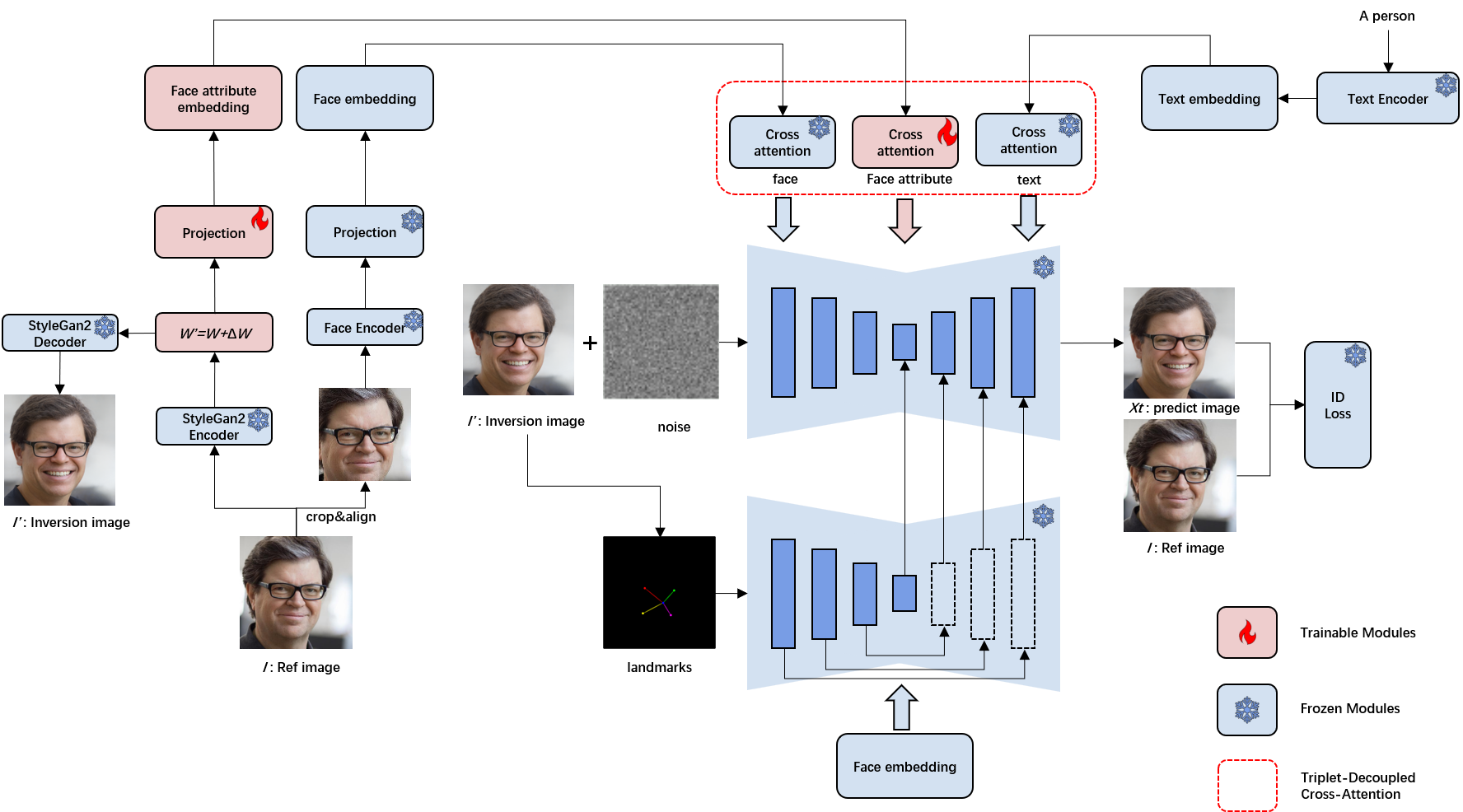}

   \caption{Overview of PSTF-AttControl framework. We use the StyleGAN2 encoder to extract facial attribute features and integrate face, attribute, and text embeddings into the diffusion model via the Triplet-Decoupled Cross-Attention module. The attribute-controlled synthesis approach for data augmentation enables the model to learn controllable facial attribute editing. We train only the projection and the Cross-Attention of facial attributes, shown in the pink modules in the figure.}
   \label{fig:architecture}
\end{figure*}

\textbf{Overview}: As shown in Figure \ref{fig:architecture}, our method builds upon and extends InstantID. We begin by using SDXL as the foundational text-to-image model. The ID encoder employs a face recognition model, while the face attribute encoder uses the e4e encoder. These are mapped to face embeddings and face attribute embeddings, respectively, via two projection modules. We then integrate these embeddings into the UNet architecture using a Triplet-Decoupled Cross-Attention mechanism, which combines face embeddings, face attribute embeddings, and text embeddings. Additionally, we incorporate a ControlNet module, conditioned on face embeddings, to provide precise facial landmark location information for the diffusion model.

\textbf{Data Augmentation with Attribute-controlled Synthesis:} 
To enable the diffusion model to appropriately respond to facial attributes in the \(W^+\) space—specifically, to adjust facial attributes by modifying the \(W^+\) space and thereby alter corresponding attributes in generated images—we incorporated Attribute-Controlled Data Augmentation Synthesis during the training process. 

\begin{itemize}
  \item The facial attribute edit directions \(\Delta W \): We utilize facial attribute edit directions from FLAME \cite{parihar2022everything} and InterfaceGAN \cite{shen2020interfacegan} to generate paired datasets consisting of edited and unedited images. The e4e encoder is then employed to extract \(W^+\) vectors for these images. By computing the mean difference between the vectors of the edited and unedited images, we obtain the corresponding edit directions in the \(W^+\) space. The face attributes used in this work include 14 categories: 'smile', 'surprise', 'angry', 'sad', 'eyesclose', 'eyeglasses', 'beard', 'gender', 'age', 'black', 'white', 'yellow', 'pose', and 'lights'.
  \item The inverted image of the edited latent: We apply the facial attribute edit directions \(\Delta W\) to the \(W^+\) space latent of the original image to perform random attribute edits. The edited latent is then converted into a new facial image using the StyleGAN2 decoder. Using the edited latent as input and the inverted image as the target, we can train the diffusion model to acquire facial attribute control capabilities. This training strategy enables the model to generate images that align with the specified attribute modifications in the latent space.  
\end{itemize}

\textbf{Triplet-Decoupled Cross-Attention}: A common approach for merging two different image embeddings is to concatenate them and pass them into a cross-attention mechanism. However, in the PSTF controllable facial attribute generation task, modifying the face attribute embedding will affect the response of the cross-attention to the face embedding. This can lead to unwanted perturbations in the facial identity. To address this, we propose Triplet-Decoupled Cross-Attention, where the face embedding and face attribute embedding are passed through independent cross-attention modules, and their outputs are weighted and summed with the text embedding. The formula for this process is as follows:

\begin{align}
    Z_{\text{new}} &= \text{Attention}(Q, K_t, V_t) + \lambda_1 \cdot \text{Attention}(Q, K_i, V_i) \nonumber \\
                  &\quad + \lambda_2 \cdot \text{Attention}(Q, K_j, V_j) \quad
    \end{align}

Here, \(Q\), \(K_t\), and \(V_t\) are the query, key, and value matrices for the text cross-attention, while \(K_i\) and \(V_i\) correspond to the face cross-attention, and \(K_j\) and \(V_j\) are for the face attribute cross-attention. The parameters \(\lambda_1\) and \(\lambda_2\) control the weight of the outputs from the face cross-attention and face attribute cross-attention, respectively.

\subsection{Training Method}
We train our model on the FFHQ facial dataset, using pre-trained weights from SDXL along with the weights from InstantID to initialize parameters, which are kept frozen during training. Only the face attribute adapter and cross-attention modules are trained, allowing for controllable facial attribute generation. The training process is structured as follows:

\textbf{Preprocessing}: For each image \( I \) in the facial dataset, we describe it with a simple text prompt \( T \) (e.g., "a person" or "portrait"). For the largest face detected in the image, we extract facial identity features \( F \), facial landmarks \( L \), and facial attribute features \( W \).

\textbf{Random attribute modification augmentation}: A facial attribute and intensity value \( \alpha \) are randomly selected. We adjust the facial attribute features as \( W' = W + \alpha \times \Delta W \). Using the StyleGAN2 decoder, we invert these modified features to generate a new facial image \( I' \) based on \( W' \).

\textbf{Training input}: We input \( (I, T, F, L, W) \) or \( (I', T, F, L, W') \) into the diffusion network for training. 

\textbf{Loss function}: Since \( I' \) may not exactly match the identity of \( I \), we include an identity loss to ensure identity consistency.

\[
L_{\text{ID}} = \left\| E_{\text{ID}}(X_t) - E_{\text{ID}}(I) \right\|_2^2
\]

\[
L = L_{\text{Diffusion}} + \lambda_{\text{ID}} L_{\text{ID}} \tag{2}
\]

For calculating \( L_{\text{ID}} \) at an intermediate denoising stage, we approximate the clean image \( X_t \) using the DDIM method and input it into the face detector. \( E_{\text{ID}} \) represents the face recognition feature extraction model.

\section{Experiments}

\subsection{Implementation Details} 

\textbf{Training Settings} Our PSTF-AttControl model is built upon the SDXL and InstantID frameworks. For identity encoding, we use Antelopev2 \cite{deng2019arcface}, which serves as the face recognition model, aligning with the approach taken by InstantID. The StyleGAN2 encoder is employed for encoding face attributes. The parameters for Triplet-Decoupled Cross-Attention are set with \(\lambda_1 = 1.0\) and \(\lambda_2 = 1.0\). The value of  \(\lambda_2\) is set to 0.5 during inference, a slight reduction from its training value of 1.0. The hyperparameter \(\lambda_2\) is set to 1.0 during training to provide a strong supervisory signal for learning attribute manipulations. For inference, however, we reduce it to 0.5, a value empirically determined to strike an optimal balance between the desired attribute edit strength and the preservation of overall image composition and harmony.

The training dataset, FFHQ, comprises 70,000 human images. We set the Random Attribute Modification rate to 0.3, with \(\alpha\) values ranging from 0 to 2.5. The ID loss weight is configured to 1.0. Images are processed at a resolution of 1024 × 1024.

We train our model on 4 NVIDIA H100 GPUs with the following configurations: a learning rate of \(1 \times 10^{-5}\), weight decay of 0.01, 100 epochs, and a batch size of 8.

\textbf{Test Settings} All results generated by our method are based on the SDXL base model, run over 50 steps with the DPM++ 2M sampler \cite{karras2022elucidating}. Following the recommended configuration, the CFG scale is set to 5.0 \cite{podell2023sdxl}. For the Triplet-Decoupled Cross-Attention, the parameters are set with \(\lambda_1 = 1.0\) and \(\lambda_2 = 0.5\).

In the facial attribute editing, we maintain the scene layout by using the same initial noise and copying the self-attention maps obtained during the generation with the unedited \( W \), following a strategy similar to local-prompt-mixing \cite{patashnik2023localizing}.

\subsection{Face Attributes Control Comparison}

\begin{figure}[t]
  \centering
   \includegraphics[width=1\linewidth]{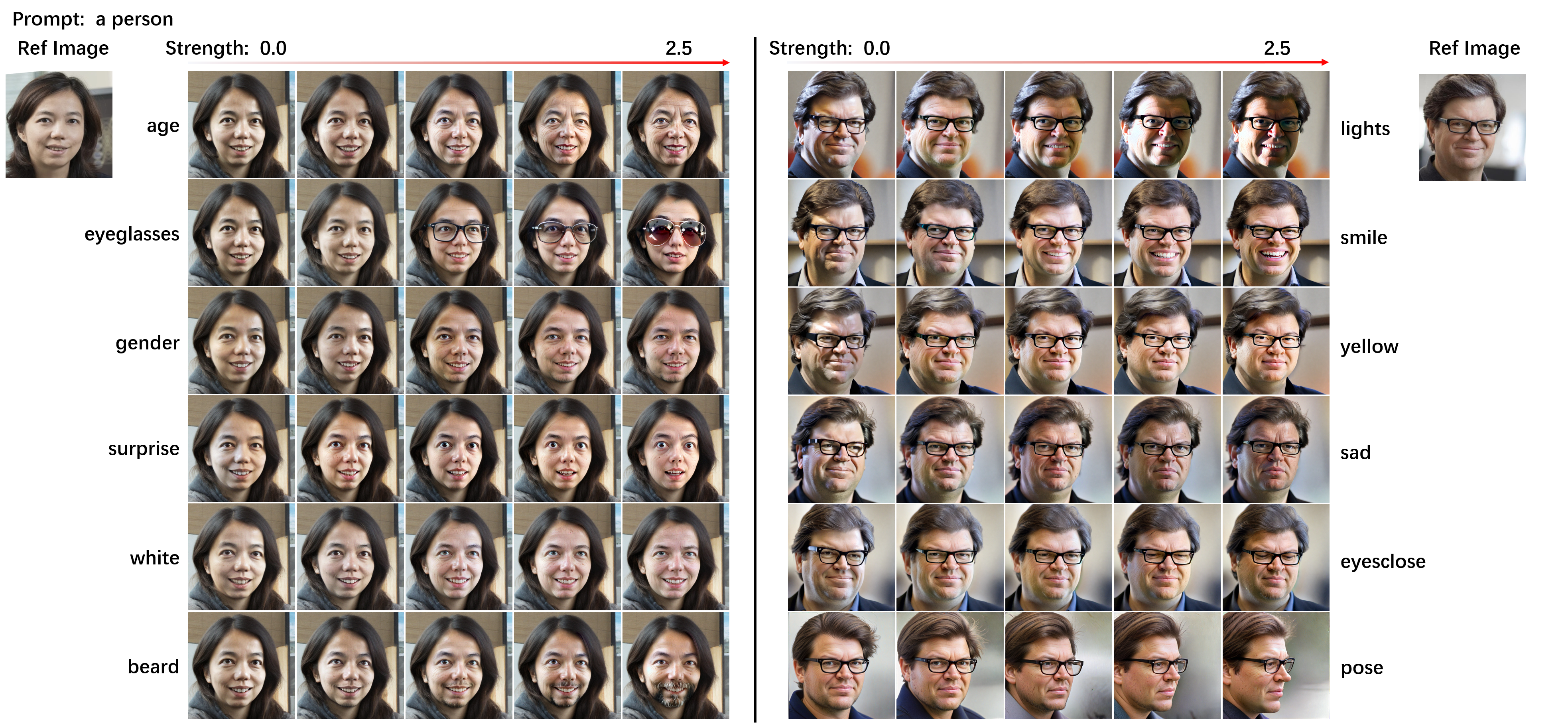}

   \caption{The Results of PSTF-AttControl. By modifying the facial attribute components in the \(W^+\) space, PSTF-AttControl enables the continuous generation of personalized images with varying attribute strengths. Here, we showcase the generation results for 12 different attributes using the faces of Fei-Fei Li and Yann LeCun as examples, demonstrating the effectiveness of our method in producing diverse, high-quality variations based on facial attributes.}
   \label{fig:attribute}
\end{figure}

Figure \ref{fig:attribute} intuitively demonstrates the superior performance of our PSTF approach in face attribute control. Given a single portrait photo and without finetuning, our method generates identity-preserving images with varying facial attributes. By adjusting the \(W^+\) latent, the attribute strength increases smoothly, leading to continuous and more pronounced changes in facial features. This precise control mechanism allows users to fine-tune facial attributes, ensuring they achieve a satisfactory result.

\textbf{Tuning-based Method Comparison:}

\begin{figure*}[t]
    \centering
     \includegraphics[width=1\linewidth]{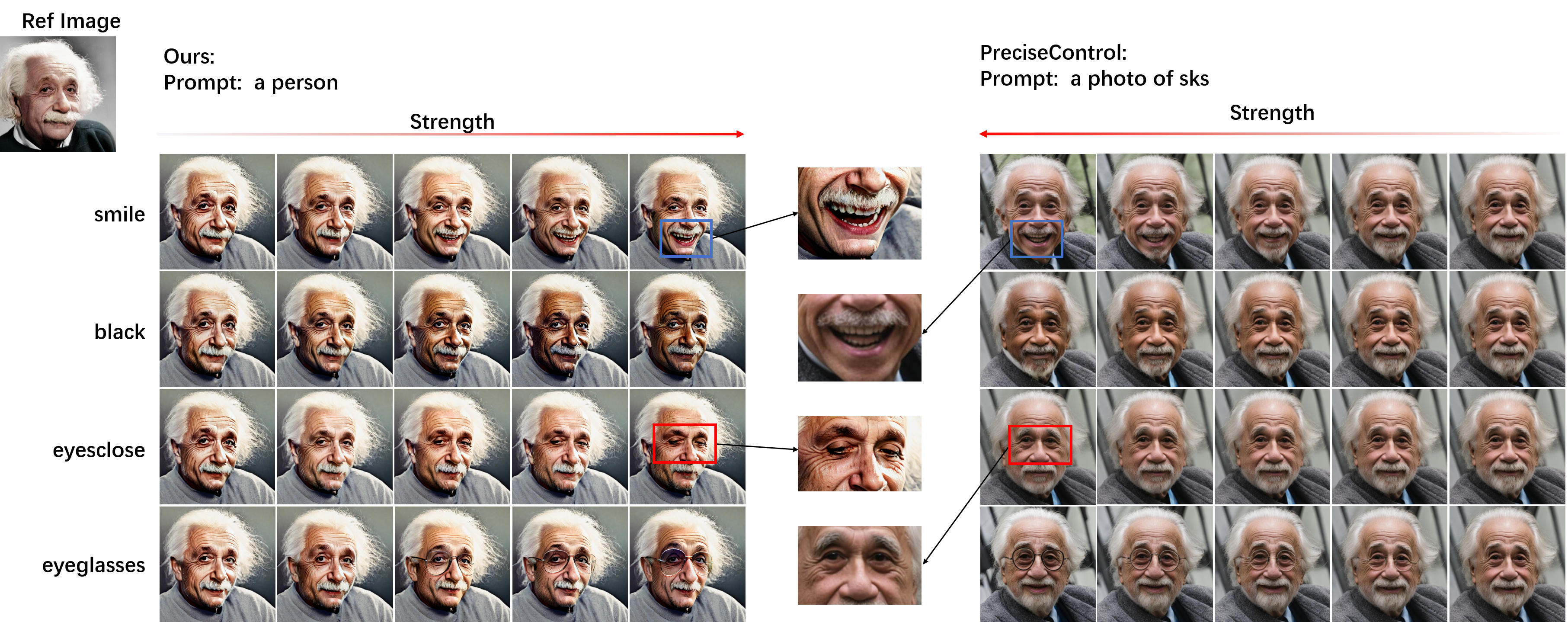}
  
     \caption{Comparison with PreciseControl. For the "smile" attribute, our method produces higher-quality teeth generation compared to PreciseControl (highlighted in blue). In the case of the "eyeclose" attribute, PreciseControl fails to make any visible changes, whereas our method smoothly closes the eyes (highlighted in red). The final column shows the mask used by PreciseControl.}
     \label{fig:precisecontrol}
  \end{figure*}

To compare with PreciseControl \cite{parihar2025precisecontrol}, multiple images of Einstein are fine-tuned using PreciseControl method to generate
a series of continuous images for each facial attribute with its default parameters. We selected four attributes—'black,' 'eyesclose,' 'smile,' and 'eyeglasses'—for demonstration. As shown in Figure \ref{fig:precisecontrol}, for the 'smile' attribute, the last image generated by PreciseControl has significant defects in the teeth, while ours shows a clean result. Furthermore, PreciseControl has no effect on the 'eyesclose' attribute, whereas our method gradually closes the eyes. PreciseControl employs a masked generation method and the promptmix technique, resulting in a higher consistency across background and subject appearances in its images compared to ours.

\textbf{PSTF Methods Comparison:}

\begin{figure*}[t]
  \centering
   \includegraphics[width=1\linewidth]{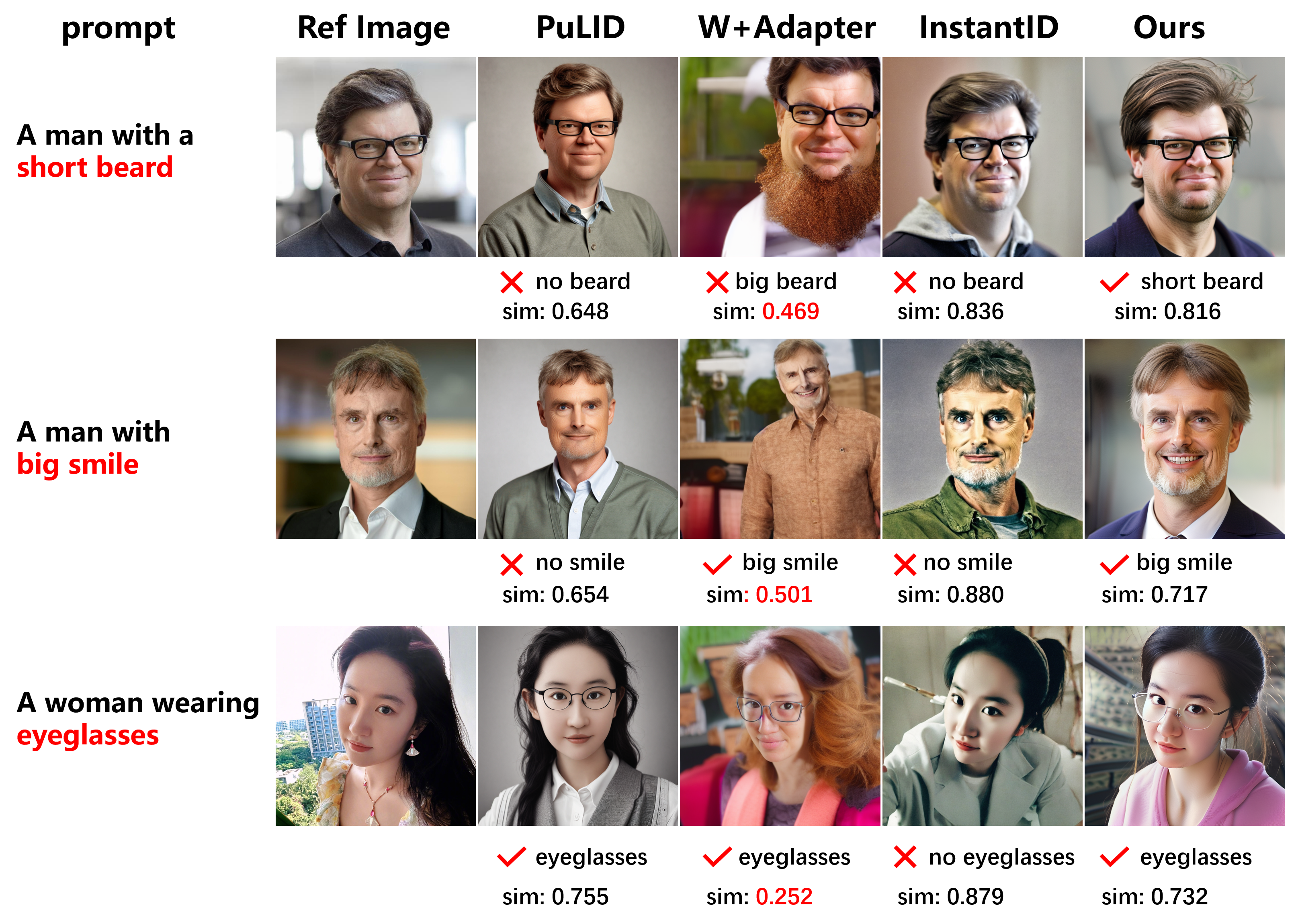}

   \caption{Comparison with PSTF Methods. The results from PuLID and InstantID show that text-based control of facial attributes is limited in its effectiveness. The results produced by W+Adapter show limited similarity to the reference face, and its manipulation of the “beard” attribute lacks precision. In contrast, our method successfully generates the desired facial features while maintaining consistent identity across faces.}
   \label{fig:compare}
\end{figure*}

As shown in Figure \ref{fig:compare}, when compared to state-of-the-art PSTF methods like InstantID, W+Adapter and PuLID, we tested three personalized generation methods for facial attributes using the following prompts: "A man with a short beard," "A man with a big smile," and "A woman wearing eyeglasses." The results generated by PuLID and InstantID indicate that text-based control of facial attributes has limited effectiveness. The results produced by W+Adapter show limited similarity to the reference face, and its manipulation of the “beard” attribute lacks precision. In contrast, our method successfully generates the desired facial features while maintaining consistent identity across faces.

\subsection{Quantitative Comparison of Facial Similarity}
\begin{table}[h!]
  \centering
  \footnotesize 
  \caption{Quantitative comparison with PSTF SOTA methods on the Unsplash-50 dataset. The results, with the "*" label, indicate that we exclude images in the PuLID method that have a facial similarity score below 0.6.}
  \label{tab:similarityComparison}

  \setlength{\tabcolsep}{8pt} 
  \begin{tabular}{l|c|c|c|c}
  \hline
  \textbf{Method} & \textbf{Cosine Similarity} & \textbf{Cosine Similarity*} & \textbf{CLIP\_T} & \textbf{CLIP\_I} \\ \hline
  PuLID           & 0.684                      & 0.757                       & \textbf{0.305}            & 0.797            \\ 
  W+Adapter       & 0.423                      & 0.62                        & 0.281            & 0.719            \\ 
  InstantID       & 0.720                      & 0.748                       & 0.291            & 0.782            \\ 
  Ours            & \textbf{0.753}             & \textbf{0.789}              & 0.289            & \textbf{0.808}            \\ \hline
  \end{tabular}
\end{table}

\begin{figure}[t]
  \centering
      \includegraphics[width=1\linewidth]{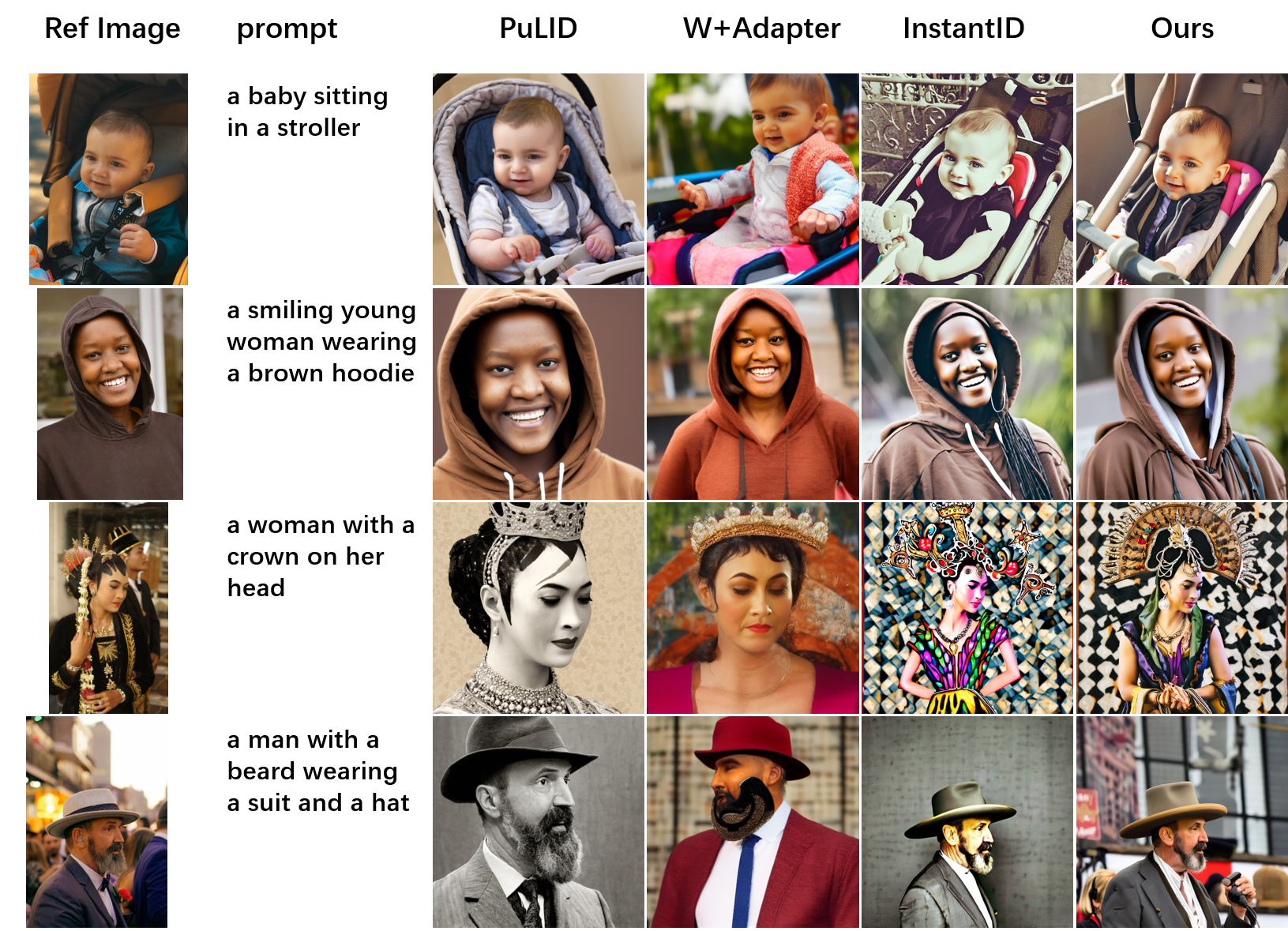}
  
      \caption{Some results of comparison with InstantID, W+Adapter and PuLID on the Unsplash-50 dataset. Our method outperformed others in preserving facial identity and fine details.}
      \label{fig:moreresults}
  \end{figure} 

To further evaluate the performance of our method in maintaining facial identity, we conducted a quantitative comparison with InstantID, W+Adapter and PuLID.

\textbf{Dataset:} We used the publicly available Unsplash-50 dataset, which consists of 50 portrait images, each paired with a corresponding caption.

\textbf{Experimental Setup:} To ensure a fair comparison, PuLID, InstantID, and our method were all implemented using the base model \texttt{SDXL-base}, which was used in the PuLID and InstantID papers. For \texttt{W+Adapter}, following its paper, we used \texttt{SD1.5} as the base model. For PuLID, we set \texttt{id\_scale} to 1.0 and \texttt{num\_zero} to 0. For InstantID and our method, we set \texttt{ip\_adapter\_scale} (corresponding to \(\lambda_1\) in our method) to 1.0 to maximize similarity. In our method, \(\lambda_2\) was set to 0.5, and we did not modify facial attributes in the \(W^+\) latent space. The random seed was fixed at 42 for all methods, and other parameters were kept at their default values. Images from Unsplash-50 were used as reference images, with the corresponding captions serving as prompts.

\textbf{Quantitative Comparison:} We utilized Antelopev2 to extract features for cosine similarity comparison, then calculated the average cosine similarity between each generated image and its corresponding reference image.

As shown in Table \ref{tab:similarityComparison}, our method outperforms W+Adapter, InstantID, and PuLID in terms of facial similarity, achieving an improvement of 0.033 over InstantID. We observed that PuLID generates many images with a small proportion of faces, which impacts its performance. Therefore, we applied the "*" label, meaning that we exclude images in the PuLID method that have a facial similarity score below 0.6.

In terms of CLIP-I and CLIP-T scores, our method achieves the highest CLIP-I score, indicating superior preservation of image–prompt consistency. For CLIP-T, our method is slightly lower than InstantID and PuLID, but remains competitive overall. 

Overall, these results demonstrate that integrating a facial attribute branch into InstantID using the StyleGAN2 encoder significantly enhances the model’s ability to preserve facial identity. As shown in Figure \ref{fig:moreresults}, our method outperforms others in maintaining facial identity and fine details, providing an intuitive visual comparison.

\subsection{Ablation Study}

\begin{figure}[t]
  \centering
      \includegraphics[width=1\linewidth]{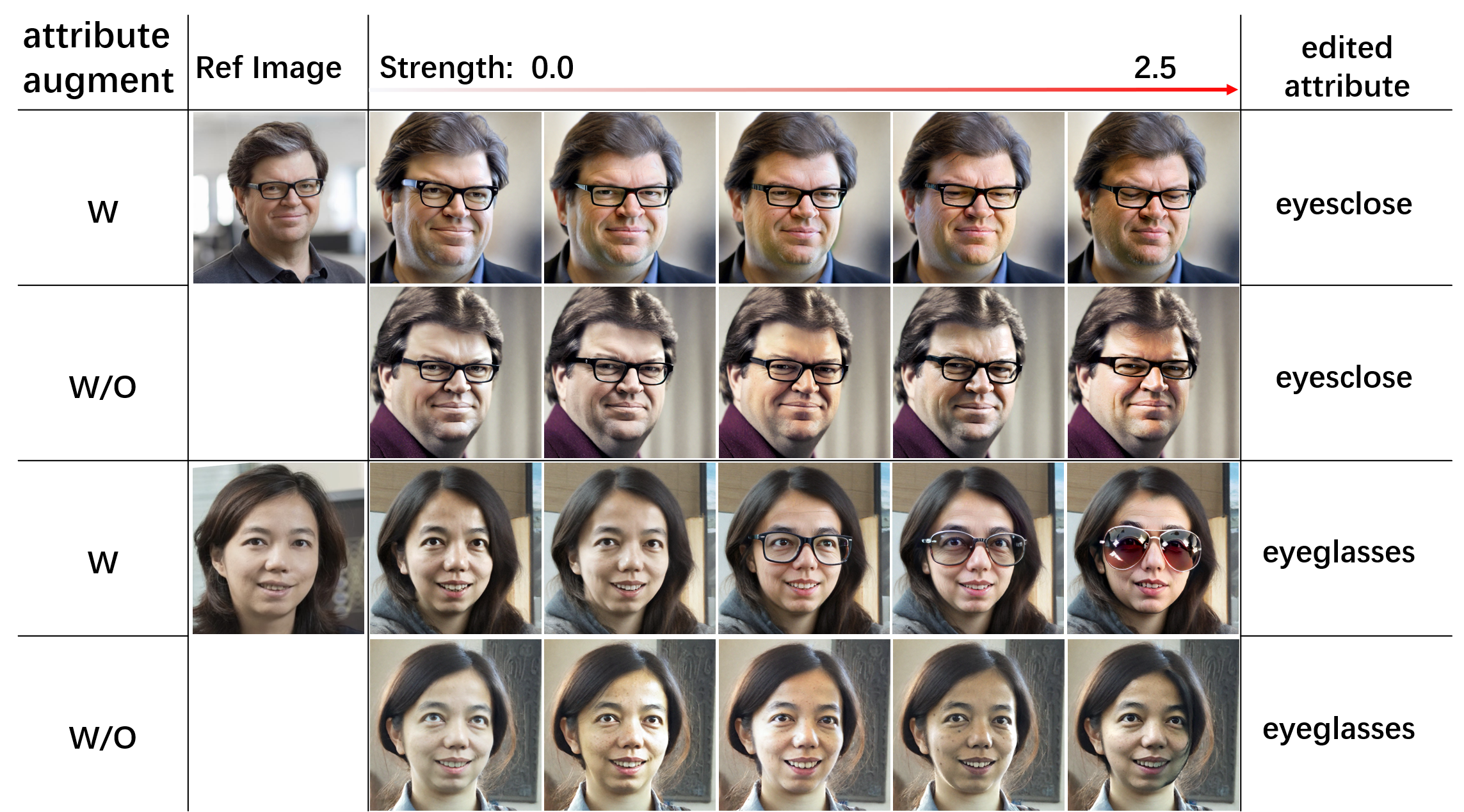}
  
      \caption{Ablation study on attribute augmentation. The facial attributes of the generated images produced by the model without attribute augmentation remained unchanged as the attribute strength increased. }
      \label{fig:woaug}
  \end{figure}

\begin{figure}[t]
    \centering
     \includegraphics[width=1\linewidth]{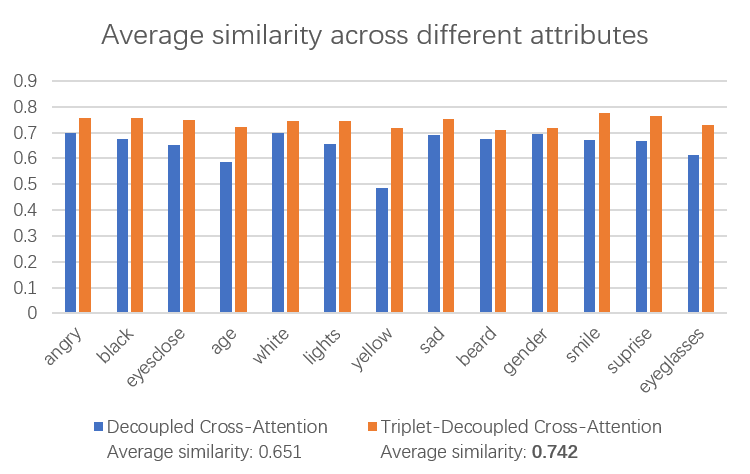}
  
     \caption{Average similarity across different attributes. The Triplet-Decoupled Cross-Attention consistently outperformed Decoupled Cross-Attention in terms of average similarity for each individual attribute.}
     \label{fig:attribute_sim}
  \end{figure}
\begin{figure}[t]
    \centering
     \includegraphics[width=1\linewidth]{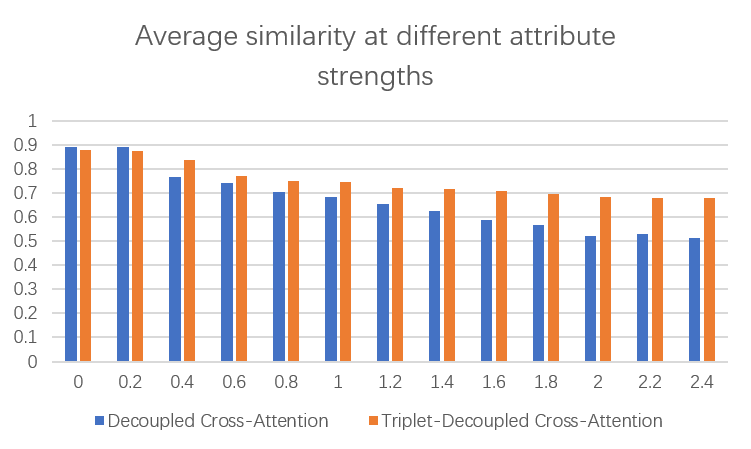}
  
     \caption{Average similarity at different attribute strengths. As attribute strength increases, average face similarity decreases for both methods. The decline is more gradual with Triplet-Decoupled Cross-Attention, which outperforms Decoupled Cross-Attention at higher attribute strengths. At attribute strengths of 0.0 and 0.2, Decoupled Cross-Attention performs slightly better.}
     \label{fig:strengths}
  \end{figure}

\textbf{Data Augmentation with Attribute-controlled Synthesis:}
To validate the necessity of Data Augmentation with Attribute-Controlled Synthesis, we compared our method with a baseline where the model was trained without attribute augmentation, while keeping all other training parameters consistent. During inference, we selected the "eyeglasses" and "eyesclose" attributes for editing. As shown in Figure \ref{fig:woaug}, the facial attributes of the generated images produced by the model without attribute augmentation remained unchanged as the attribute strength increased, demonstrating the importance of attribute augmentation for enabling controllable attribute manipulation.

\textbf{Triplet-Decoupled Cross-Attention:}
To validate the effectiveness of the Triplet-Decoupled Cross-Attention, we conducted the following comparative experiments.

Decoupled Cross-Attention: We concatenated face embeddings and face attribute embeddings, then used the Decoupled Cross-Attention structure to inject the concatenated embedding and text embedding into the U-Net. During model training, we initialized the model using the parameters from SDXL and InstantID, just as in the original setup. We only trained the projection module of the face attribute feature branch and the cross-attention module for the concatenated embedding. All other parameters were identical to those used in the experiments of our proposed method.

To evaluate the performance of the two methods, we compared the similarity between the generated face images and the reference face images. We collected 20 face images from the internet (excluding those from the FFHQ dataset) and used both methods to generate portrait images. For each reference image, we generated 169 images by varying all facial attributes with parameter \(\alpha\) values ranging from 0 to 2.5 in increments of 0.2. We used Antelopev2 \cite{deng2019arcface} to extract features for cosine similarity comparison.

As shown in Figure \ref{fig:attribute_sim}, the Triplet-Decoupled Cross-Attention structure improved the average face similarity across all images by 0.091 compared to the Decoupled Cross-Attention method. Furthermore, TDCA outperformed Decoupled Cross-Attention in terms of average similarity for each individual attribute. Figure \ref{fig:strengths} shows the average similarity of the two methods at different attribute strengths. As the attribute strength increased, the average face similarity decreased for both methods. However, the decline in similarity was much more gradual with Triplet-Decoupled Cross-Attention compared to Decoupled Cross-Attention. At attribute strengths of 0.0 and 0.2, Decoupled Cross-Attention slightly outperformed Triplet-Decoupled Cross-Attention. We attribute this to the freezing of the cross-attention in the face feature branch in our method, though we consider this slight difference negligible.

\section{Conclusions}
We proposed a novel PSTF approach, PSTF-AttControl, that enables precise control over facial attribute generation while preserving high-fidelity facial identity. Our method outperforms tune-based methods such as PreciseControl, as well as PSTF state-of-the-art approaches like InstantID, W+Adapter and PuLID, in terms of facial attribute control.

In our approach, we introduce the StyleGAN2 encoder as the facial attribute feature extraction module. By combining this with attribute-controlled synthesis data augmentation, the model learns the ability to perform controllable facial attribute editing. Furthermore, by utilizing Triplet-Decoupled Cross-Attention, we significantly enhance the face similarity between the generated images and the reference images, especially when modifying facial attributes.

This work demonstrates the potential of PSTF methods in achieving high-quality, controllable facial attribute editing while maintaining the integrity of the original identity, offering a promising direction for future advancements in personalized image generation.




\bibliographystyle{elsarticle-num.bst}
\bibliography{lxbibfile}

%% For numbered reference style

%   \textit{\LaTeX: a document preparation system},
%   Addison Wesley, Massachusetts,
\end{document}